%% file: eacl2021.tex
\documentclass[11pt,a4paper]{article}
\usepackage[hyperref]{eacl2021}
\usepackage{tikz}
\usepackage{times}
\usepackage{latexsym, amsmath, amssymb}
\usepackage{amsfonts}
\usepackage{multirow}
\usepackage{graphicx}

\usepackage{soul}

\usepackage{circuitikz}
\usepackage{dsfont}
\usepackage{commath}

\usepackage[utf8]{inputenc}
\usepackage{pgfplots}
\DeclareUnicodeCharacter{2212}{−}
\usepgfplotslibrary{groupplots,dateplot}
\usetikzlibrary{patterns,shapes.arrows}
\pgfplotsset{compat=newest}

\usepackage{placeins}
\usepackage{subcaption}
\usepackage{mwe}
\usepackage{pifont}
\usepackage{xcolor}
\usepackage{soul}
\usepackage{paralist}

\usepackage{booktabs}
\usepackage{relsize}
\usepackage{array}
\newcolumntype{V}{>{\smaller}l}

\colorlet{Mycolor1}{green!10!orange!90!}
\newcommand{\final}[1]{{#1}}

\usepackage{xspace}
\usepackage{adjustbox}

\newcommand{\tabref}[1]{Table~\ref{table:#1}\xspace}

\newcommand{\adv}{\ensuremath{\text{adv}}}
\newcommand{\diff}{\ensuremath{\text{diff}}}
\newcommand{\linear}{\ensuremath{\text{linear}}}
\newcommand{\z}{\phantom{0}}

\definecolor{acc0}{rgb}{0.12156862745098,0.466666666666667,0.705882352941177}
\definecolor{acc1}{rgb}{1,0.498039215686275,0.0549019607843137}
\definecolor{tpr_gap}{rgb}{0.172549019607843,0.627450980392157,0.172549019607843}
\definecolor{tnr_gap}{rgb}{0.83921568627451,0.152941176470588,0.156862745098039}

\usepackage{mathtools, caption}
\DeclareRobustCommand\sampleline[1]{%
  \tikz\draw[semithick,#1] (0,0) (0,\the\dimexpr\fontdimen22\textfont2\relax)
  -- (2em,\the\dimexpr\fontdimen22\textfont2\relax);%
}

\usepackage{microtype}

\aclfinalcopy % Uncomment this line for the final submission
\title{Diverse Adversaries for Mitigating Bias in Training}

\author{
  Xudong Han \\\And
  Timothy Baldwin \\
  School of Computing and Information Systems \\
  The University of Melbourne \\
  Victoria 3010, Australia \\
  \texttt{xudongh1@student.unimelb.edu.au} \\
  \texttt{\{tbaldwin,tcohn\}@unimelb.edu.au} \\\And
  Trevor Cohn \\
}

\date{}
\begin{document}
\maketitle

\begin{abstract}
Adversarial learning can learn fairer and less biased models of language than standard methods.
However, current adversarial techniques only partially mitigate model bias, added to which their training procedures are often unstable. 
In this paper, we propose a novel approach to adversarial learning based on the use of multiple \emph{diverse} discriminators, 
whereby discriminators are encouraged to learn orthogonal hidden representations from one another.
Experimental results show that our method substantially improves over
standard adversarial removal methods, in terms of reducing bias and the stability of training.
\end{abstract}

\section{Introduction}

While NLP models have achieved great successes, results can depend on spurious correlations with protected attributes of the authors of a given text, such as gender, age, or race. 
Including protected attributes in models can lead to problems such as
leakage of personally-identifying information of the author
%elazar-goldberg-2018-adversarial, 
\citep{li-etal-2018-towards}, and
unfair models, i.e., models which do not perform equally well for
different sub-classes of user. 
This kind of unfairness has been shown to exist in many different tasks, including part-of-speech tagging \citep{hovy-sogaard-2015-tagging} and sentiment analysis \citep{kiritchenko-mohammad-2018-examining}.

One approach to diminishing the influence of protected attributes is to use adversarial methods, where an encoder attempts to prevent a discriminator from identifying the protected attributes in a given task %elazar-goldberg-2018-adversarial, 
\citep{li-etal-2018-towards}. 
Specifically, an adversarial network is made up of an attacker and encoder, where the attacker detects protected information in the representation of the encoder, and the optimization of the encoder incorporates two parts: (1) minimizing the main loss, and (2) maximizing the attacker loss (i.e., preventing protected attributes from being detected by the attacker).
Preventing protected attributes from being detected tends to result in fairer models, as protected attributes will more likely be independent rather than confounding variables.
Although this method leads to demonstrably less biased models, there are still limitations,
%The main problem is that 
most notably that significant protected information still remains in the model's encodings and prediction outputs \citep{wang2019balanced, elazar-goldberg-2018-adversarial}.

Many different approaches have been proposed to strengthen the attacker, including: increasing the discriminator hidden dimensionality; assigning different weights to the adversarial component during training; using an ensemble of adversaries with different initializations; and reinitializing the adversarial weights every $t$ epochs  \citep{elazar-goldberg-2018-adversarial}. Of these, the ensemble method has been shown to perform best, but independently-trained attackers can generally still detect private information after adversarial removal.

In this paper, we adopt adversarial debiasing approaches and present a novel way of strengthening the adversarial component via orthogonality constraints \citep{salzmann2010factorized}.
Over a sentiment analysis dataset with racial labels of the document
authors, we show our method to result in both more accurate and fairer
models, with privacy leakage close to the lower-bound.\footnote{Source code available at \url{https://github.com/HanXudong/Diverse_Adversaries_for_Mitigating_Bias_in_Training}}

\section{Methodology}

Formally, given an input $x_{i}$ annotated with main task label $y_{i}$ and protected attribute label $g_{i}$, a main task model $M$ is trained to predict $\hat{y}_{i}=M(x_{i})$, and an adversary, aka ``discriminator'', $A$ is trained to predict $\hat{g}_{i}=A(h_{M,i})$ from $M${'}s last hidden layer representation $h_{M,i}$. 
In this paper, we treat a neural network classifier as a combination of two connected parts: (1) an encoder $E$, and (2) a linear classifier $C$. 
For example, in the main task model $M$, the encoder $E_{M}$ is used to compute the hidden representation $h_{M,i}$ from an input $x_{i}$, i.e., $h_{M,i}=E_{M}(x_{i})$, and the decoder is used to make a prediction, $\hat{y}_{i}=C_{M}(h_{M,i})$. 
Similarly, for a discriminator, $\hat{g}_{i}=A(h_{M,i})=C_{A}(E_{A}(h_{M,i}))$.

\subsection{Adversarial Learning}
Following the setup of \citet{li-etal-2018-towards} and \citet{elazar-goldberg-2018-adversarial} the optimisation objective for our standard adversarial training is:
% \begin{equation}
%     \label{eq:adv_optimization}
    $$\min_{M}\max_{A}\mathcal{X}(y, \hat{y}_M) - \lambda_{\adv} \mathcal{X}(g, \hat{g}_A),$$
% \end{equation}
where $\mathcal{X}$ is cross entropy loss, and $\lambda_{\adv}$ is the trade-off hyperparameter. 
Solving this minimax optimization problem encourages the main task model hidden representation $h_{M}$ to be informative to $C_{M}$ and uninformative to $A$.
%When solving this optimization problem, \trevor{huh? $A$ is not updated} and the negative sign \trevor{its really the min/max, not the negative summand} is implemented as a gradient-reversal layer \citep{ganin2015unsupervised}.
Following \citet{ganin2015unsupervised}, the above can be trained using stochastic gradient optimization with a gradient reversal layer for $\mathcal{X}(g, \hat{g}_A)$.

\subsection{Differentiated Adversarial Ensemble} 

\begin{figure}
\centering

\tikzstyle{block} = [draw, fill=blue!20, rectangle, rounded corners,
    minimum height=0.75cm, minimum width=0.75cm]
\tikzstyle{sum} = [draw, fill=green!20, circle, node distance=2cm]
\tikzstyle{input} = [coordinate]
\tikzstyle{output} = [coordinate]
\tikzstyle{pinstyle} = [pin edge={to-,thin,black}]
\tikzstyle{ArrowC1} = [rounded corners, thick]

\usetikzlibrary{decorations,decorations.pathmorphing,decorations.pathreplacing}

% The block diagram code is probably more verbose than necessary
\begin{circuitikz}[auto, node distance=1cm,>=latex]
    % We start by placing the blocks
    % \node [input, name=input] {};
    \node [block, name=input] {$E_{M}$};
    \node [input, left of=input, node distance=1cm] (input_x) {};
    % \node [block, below of=input] (Encoder) {Encoder}
    \node [input, right of=input, node distance=1.5cm] (hidden) {};
    \node [block, right of=hidden, node distance=1.5cm] (adv1) {$E_{A_{1}}$};
    \node [block, below of=adv1, node distance=1.75cm] (adv2) {$E_{A_{k}}$};
    \node [block, above of=adv1,fill=green!20, node distance=1.75cm] (MC) {$C_{M}$};
    
    \node [block, fill=green!20, right of=adv1, node distance=2cm] (dense1) {$C_{A_{1}}$};
    \node [block, fill=green!20, right of=adv2, node distance=2cm] (dense2) {$C_{A_{k}}$};
    
    \node [output, right of=dense1, node distance=1.5cm] (out1) {};
    \node [output, right of=dense2, node distance=1.5cm] (out2) {};
    \node [output, right of=MC, node distance=1.5cm] (out0) {};
    
    \draw [-] (input) -- node[] {$h_{M,i}$} (hidden);
    \draw [->] (input_x) -- node[] {$x_{i}$} (input);
    \draw[->, dashed] (hidden) -- (adv1);
    \draw[->, rounded corners=.15cm, dashed] (hidden.south) |- (adv2.west);
    \draw[->, rounded corners=.15cm] (hidden.south) |- (MC.west);
    
    \path (adv1.south)+(0.0,-0.4) node [] {$\vdots$}; 
    \path (dense1.south)+(0.0,-0.4) node [] {$\vdots$}; 
    
    \draw [->] (adv1) -- node[] {$h_{A_{1},i}$} (dense1);
    \draw [->] (adv2) -- node[] {$h_{A_{k},i}$} (dense2);
    
    \draw [->] (dense1) -- node[] {$\hat{g}_{A_{1},i}$} (out1);
    \draw [->] (dense2) -- node[] {$\hat{g}_{A_{k},i}$} (out2);
    \draw [->] (MC) -- node[] {$\hat{y}_{i}$} (out0);
        
\end{circuitikz}
\caption{
\label{fig:separation_adv}
Ensemble adversarial method. Dashed lines denote gradient reversal in
adversarial learning.  The $k$ sub-discriminators $A_i$ are independently initialized.
Given a single input $x_{i}$, the main task encoder computes a hidden
representation $h_{M,i}$, which is used as the input to the main model output layer and sub-discriminators. 
From the $k$-th sub-discriminator, the estimated protected attribute label is $\hat{g}_{A_{k},i}=C_{A_{k}}(E_{A_{k}}(h_{M,i}))$. 
}
\end{figure}
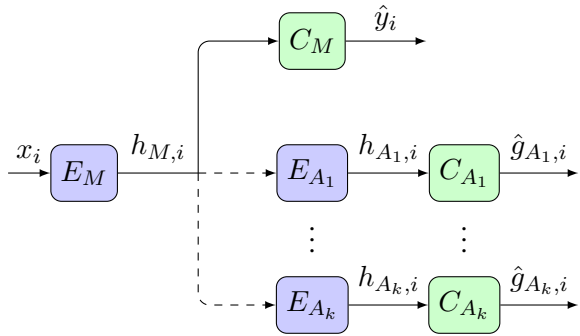

Inspired by the ensemble adversarial method \cite{elazar-goldberg-2018-adversarial} and domain separation networks \cite{bousmalis2016domain}, we present \emph{differentiated adversarial ensemble}, a novel means of strengthening the adversarial component. 
Figure~\ref{fig:separation_adv} shows a typical ensemble architecture where $k$ sub-discriminators are included in the adversarial component, leading to an averaged adversarial regularisation term:
$$-\frac{\lambda_{\adv}}{k}\sum_{j\in\{1,...,k\}} \mathcal{X}(g, \hat{g}_{A_{j}}).$$

One problem associated with this ensemble architecture is that it cannot ensure that different sub-discriminators focus on different aspects of the representation. 
Indeed, experiments have shown that sub-discriminator ensembles can weaken the adversarial component \cite{elazar-goldberg-2018-adversarial}.
To address this problem, we further introduce a difference loss \cite{bousmalis2016domain} to encourage the adversarial encoders to encode different aspects of the private information. 
As can be seen in Figure~\ref{fig:separation_adv}, $h_{A_{k},i}$ denotes the output from the $k$-th sub-discriminator encoder given a hidden representation $h_{M,i}$, i.e., $h_{A_{k},i}=E_{A_{k}}(h_{M,i})$. 

The difference loss encourages orthogonality between the encoding representations of each pair of sub-discriminators: 
% \begin{equation}
    % \label{eq:differenceLoss}
    $$\mathcal{L}_{\diff} = \lambda_{\diff}\sum_{i,j\in\{1,...,k\}}\norm{{h_{A_{i}}}^{\intercal}h_{A_{j}}}_{F}^{2}\mathds{1}(i\neq{j}),$$
% \end{equation}
where $\norm{\cdot}_{F}^{2}$ is the squared Frobenius norm.

Intuitively, sub-discriminator encoders must learn different ways of
identifying protected information given the same input embeddings,
resulting in less biased models than the standard ensemble-based adversarial method.
\final{
According to \citet{bousmalis2016domain}, the difference loss has the additional advantage of also being minimized when hidden representations shrink to zero. Therefore, instead of minimizing the difference loss by learning rotated hidden representations (i.e., the same model), this method biases adversaries to have representations that are a) orthogonal, and b) low magnitude; the degree to which is given by weight decay of the optimization function.
}

%  the table for the AAE vs SAE dataset
\begin{table*}[!t]
\centering
\begin{adjustbox}{max width=\linewidth}
\begin{tabular}{ll@{\smaller$\pm$}Vr@{\smaller$\pm$}Vr@{\smaller$\pm$}Vl@{\smaller$\pm$}Vl@{\smaller$\pm$}V}
\toprule
Model         & \multicolumn{2}{c}{Accuracy$\uparrow$}     &\multicolumn{2}{c}{TPR Gap$\downarrow$}&\multicolumn{2}{c}{TNR Gap$\downarrow$}&\multicolumn{2}{c}{Leakage@$\mathbf{h}$$\downarrow$}& \multicolumn{2}{c}{Leakage@$\hat{y}$$\downarrow$} \\ 
\midrule
Random        & 50.00    &     0.00              & 0.00   &  0.00            & 0.00   &  0.00            & \multicolumn{2}{c}{---\z\z}   & \multicolumn{2}{c}{---\z\z} \\
  Fixed Encoder & 61.44    &     0.00               & 0.52   &  0.00             & 17.97  &  0.00             & 92.07  &  0.00            & 86.93  &  0.00\\[1ex]

Standard      & 71.59    &     0.05              & 31.81  &  0.29            & 48.41  &  0.27            & 85.56  &  0.20            & 70.09  &  0.19 \\[1ex]

% INLP        & 68.54     & 1.05      & 25.13     & 2.31      & 40.70     & 5.02  & \bf 66.64     & 0.87 & \multicolumn{2}{c}{---\z\z} \\

INLP        & 68.54     & 1.05      & 25.13     & 2.31      & 40.70     & 5.02  & \bf 66.64     & 0.87 & 66.19 & 0.79 \\

Adv Single Discriminator           & 74.25    &     0.39              & 13.01  &  3.83            & 28.55  &  3.60            & 84.33  &  0.98            & 61.48  &  2.17 \\
% INLP          & 62.98    &     3.18              & 19.06  &  4.81            & 29.92  &  11.19           &\bf59.18&  9.50            &\bf57.99&  13.81 \\
Adv Ensemble      & 74.08    &     0.99              & 12.04  &  3.50            & 31.76  &  3.19            & 85.31  &  0.51            & 63.23  &  3.62 \\[1ex]

Differentiated Adv Ensemble  & \bf 74.52&     0.28              &\bf 8.42&  1.84            & \bf 24.74  &  2.07            & 84.52  &  0.50            & \bf 61.09  &  2.32 \\
% ADV+INLP      & 70.93    &     2.07              &  9.33  &  3.66            &\bf15.67&  8.44            & 83.79  &  0.90            & 59.37  &  4.08 \\
\bottomrule
\end{tabular}
\end{adjustbox}
\caption{
Evaluation results $\pm$ standard deviation ($\%$) on the test set, averaged over 10 runs with different random seeds. 
\textbf{Bold} = best performance. ``$\uparrow$'' and ''$\downarrow$'' indicate that higher and lower performance, resp., is better for the given metric. Leakage measures the accuracy of predicting the protected attribute, over the final hidden representation $\mathbf{h}$ or model output $\hat{y}$. Since the Fixed Encoder is not designed for binary sentiment classification, we merge the original 64 labels into two categories based on the results of hierarchical clustering.}
\label{table:results}
\end{table*}

\subsection{INLP}
\final{
We include Iterative Null-space Projection (``INLP'': \citet{ravfogel-etal-2020-null}) as a baseline method for mitigating bias in trained models, in addition to standard and ensemble adversarial methods.
In INLP, a linear discriminator ($A_{\linear}$) of the protected attribute is iteratively trained from pre-computed fixed hidden representations (i.e., $h_{M}$) to project them onto the linear discriminator's null-space, $h_{M}^{*} = P_{N(A_{\linear})}h_{M}$, where $P_{N(A_{\linear})}$ is the null-space projection matrix of $A_{\linear}$. In doing so, it becomes difficult for the protected attribute to be linearly identified from the projected hidden representations ($h_{M}^{*}$), and any linear main-task classifier ($C_{M}^{*}$) trained on $h_{M}^{*}$ can thus be expected to make fairer predictions.
}

\section{Experiments}
\paragraph{Fixed Encoder}
\final{
Following \citet{elazar-goldberg-2018-adversarial} and \citet{ravfogel-etal-2020-null}, we use the DeepMoji model \citep{felbo-etal-2017-using} as a fixed-parameter encoder (i.e.\ it is not updated during training). The DeepMoji model is trained over 1246 million tweets containing one of 64 common emojis. We merge the 64 emoji labels output by DeepMoji into two super-classes based on hierarchical clustering: `happy' and `sad'.
}

\paragraph{Models}
The encoder $E_{M}$ consists of a fixed pretrained encoder (DeepMoji) and two trainable fully connected layers (``Standard'' in \tabref{results}).
Every linear classifier ($C$) is implemented as a dense layer.

For protected attribute prediction, a discriminator ($A$) is a 3-layer MLP where the first 2 layers are collectively denoted as $E_{A}$, and the output layer is denoted as $C_{A}$. 

\paragraph{TPR-GAP and TNR-GAP}
In classification problems, a common way of measuring bias is TPR-GAP and TNR-GAP, which evaluate the gap in the True Positive Rate (TPR) and True Negative Rate (TNR), respectively, across different protected attributes \citep{10.1145/3287560.3287572}. 
This measurement is related to the criterion that the prediction $\hat{y}$ is conditionally independent of the protected attribute $g$ given the main task label $y$ (i.e., $\hat{y} \bot g | y$). 
Assuming a binary protected attribute, this conditional independence requires $\mathbb{P}\{\hat{y}|y, g=0\} = \mathbb{P}\{\hat{y}|y, g=1\}$, which implies an objective that minimizes the difference (GAP) between the two sides of the equation.

\paragraph{Linear Leakage}
We also measure the leakage of protected attributes. A model is said to leak information if the protected attribute can be predicted at a higher accuracy than chance, in our case, from the hidden representations the fixed encoder generates.
We empirically quantify leakage with a linear support vector classifier at two different levels: 
\begin{compactitem}
    \item {Leakage@$\mathbf{h}$}: the accuracy of recovering the protected attribute from the output of the final hidden layer after the activation function ($h_{M}$). 
    \item {Leakage@$\hat{y}$}: the accuracy of recovering the protected attribute from the output $\hat{y}$ (i.e., the logits) of the main model.
\end{compactitem}
% Recall that the output layer of the main model ($C_{M}$) is assumed to be a linear classifier, and thus we only need to detect linear leakage from the input of $C_{M}$ which is $h_{M}$.

% \subsection{Race information in tweets}

\paragraph{Data}
\final{ 
We experiment with the dataset of \citet{blodgett-etal-2016-demographic}, which contains tweets that are either African American English (AAE)-like or Standard American English (SAE)-like (following \citet{elazar-goldberg-2018-adversarial} and \citet{ravfogel-etal-2020-null}).
}
Each tweet is annotated with a binary ``race'' label (on the basis of AAE or SAE) and a binary sentiment score, which is determined by the (redacted) emoji within it.

In total, the dataset contains 200k instances, perfectly balanced across the four race--sentiment combinations.
To create bias in the dataset, we follow previous work in skewing the training data to generate race--sentiment combinations (AAE--happy, SAE--happy, AAE--sad, and SAE--sad) of $40\%$, $10\%$, $10\%$, and $40\%$, respectively.
Note that we keep the test data unbiased.

\begin{figure}[t!]
    \centering
    \pgfplotsset{width=0.48\textwidth, height = 4cm, compat=1.16}
    \pgfplotsset{every axis/.append style={
                    label style={font=\small},
                    tick label style={font=\scriptsize}  
                    }}
    \input{Figures/figure2}
    \caption{$\lambda_{\adv}$ sensitivity analysis, averaged over 10 runs for a single discriminator adversarial model. Main task accuracy of group \textcolor{acc0}{\textit{SAE}} (blue) and \textcolor{acc1}{\textit{AAE}} (orange), \textcolor{tpr_gap}{\textit{TPR-GAP}} (green), and \textcolor{tnr_gap}{\textit{TNR-GAP}} (red) are reported.
    } 
    
    \label{fig:original_adv}
\end{figure}
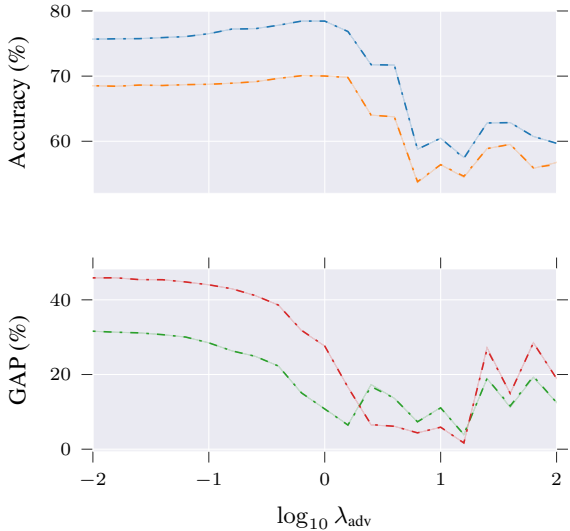

\paragraph{Training Details} 
All models are trained and evaluated on the same training/test split.
The Adam optimizer \citep{kingma:adam} is used with learning rates of $3\times10^{-5}$ for the main model and $3\times10^{-6}$ for the sub-discriminators. The minibatch size is set to 1024. 
Sentence representations (2304d) are extracted from the DeepMoji encoder.
The hidden size of each dense layer is 300 in the main model, and 256 in the sub-discriminators.
We train $M$ for 60 epochs and each $A$ for 100 epochs, keeping the checkpoint model that performs best on the dev set.
\final{
Similar to \citet{elazar-goldberg-2018-adversarial}, hyperparameters ($\lambda_{\adv}$ and $\lambda_{\diff}$) are tuned separately rather than jointly.
$\lambda_{\adv}$ is tuned to $0.8$ based on the standard (single-discriminator) adversarial learning method, and this setting is used for all other adversarial methods. When tuning $\lambda_{\adv}$, we considered both overall performance and bias gap (both over the dev data).  Since adversarial training can increase overall performance while decreasing the bias gap (see Figure~\ref{fig:original_adv}), we select the adversarial model that achieves the best task performance.
For adversarial ensemble and differentiated models, we tune the hyperparameters (number of sub attackers and $\lambda_{\diff}$) to achieve a similar bias level while getting the best overall performance. 
To compare with a baseline ensemble method with a similar number of parameters, we also report results for an adversarial ensemble model with 3 sub-discriminators.
The scalar hyperparameter of the difference loss ($\lambda_{\diff}$) is tuned through grid search from $10^{-4}$ to $10^{4}$, and set to $10^{3.7}$.
For the INLP experiments, fixed sentence representations are extracted from the same data split. Following \citet{ravfogel-etal-2020-null}, in the INLP experiments, both the discriminator and the classifier are implemented in scikit-learn as linear SVM classifiers \citep{scikit-learn}. 
We report Leakage@$\hat{y}$ for INLP based on the predicted confidence scores, which could be interpreted as logits, of the linear SVM classifiers.
}

\paragraph{Results and Analysis}
\tabref{results} shows the results over the test set. 
Training on a biased dataset without any fairness restrictions leads to a biased model, as seen in the Gap and Leakage results for the Standard model.
\final{
Consistent with the findings of \citet{ravfogel-etal-2020-null}, INLP can only reduce bias at the expense of overall performance. On the other hand, 
}
the Single Discriminator and Adv(ersarial) Ensemble baselines both enhance accuracy and reduce bias, consistent with the findings of \citet{li-etal-2018-towards}. 

Compared to the Adv Ensemble baseline, incorporating the difference loss in our method has two main benefits: training is more stable (results have smaller standard deviation), and there is less bias (the TPR and TNR Gap are smaller). 
Without the orthogonality factor, $\mathcal{L}_{\diff}$, the sub-discriminators tend to learn similar representations, and the ensemble degenerates to a standard adversarial model.
Simply relying on random initialization to ensure sub-discriminator diversity, as is done in the Adv Ensemble method, is insufficient.
The orthogonality regularization in our method leads to more stable and overall better results in terms of both accuracy and TPR/TNR Gap.

As shown in \tabref{results}, even the Fixed Encoder model leaks protected information, as a result of implicit biases during pre-training. 
\final{
INLP achieves significant improvement in terms of reducing linear hidden representation leakage. The reason is that {Leakage@$\mathbf{h}$} is directly correlated with the objective of INLP, in minimizing the linear predictability of the protected attribute from the $\mathbf{h}$.
}
Adversarial methods do little to mitigate {Leakage@$\mathbf{h}$}, but substantially decrease {Leakage@$\hat{y}$} in the model output.
However, both types of leakage are well above the ideal value of 50\%, and therefore none of these methods can be considered as providing meaningful privacy, in part because of the fixed encoder.
This finding implies that when applying adversarial learning, the pretrained model needs to be fine-tuned with the adversarial loss to have any chance of generating a truly unbiased hidden representation.
% the original assumption has been removed
% I don't understand: mitigate protected information in a non-linear way. 
Despite this, adversarial training does reduce the TPR and TNR Gap, and improves overall accuracy, which illustrates the utility of the method for both bias mitigation and as a form of regularisation. %\footnote{Debiasing encourages the model not to use features of the input that, in many cases, should not have a major impact on the end task, and therefore debiasing has the effect of forcing the model to learn the language phenomenon underlying the task.}

Overall, our proposed method empirically outperforms the baseline models in terms of debiasing, with a better performance--fairness trade-off. 

\paragraph{Robustness to $\lambda_{\adv}$}

% \begin{figure}[t!]
%     \centering
%     \includegraphics[trim={0 2cm 0cm 0},clip,width=0.35\textwidth]{Figures/original_adv_v3.pdf}
%     \caption{$\lambda_{\adv}$ sensitivity analysis, averaged over 10 runs for a single discriminator adversarial model.}
%     \label{fig:original_adv}
% \end{figure}

We first evaluate the influence of the trade-off hyperparameter $\lambda_{\adv}$ in adversarial learning.
As can be seen from Figure~\ref{fig:original_adv}, $\lambda_{\adv}$ controls the performance--fairness trade-off.
Increasing $\lambda_{\adv}$ from $10^{-2}$ to around $10^{-0}$, TPR Gap and TNR Gap consistently decrease, while the accuracy of each group rises.
To balance up accuracy and fairness, we set $\lambda_{\adv}$ to $10^{-0.1}$.
We also observe that an overly large $\lambda_{\adv}$ can lead to a more biased model (starting from about $10^{1.2}$).

\paragraph{Robustness to $\lambda_{\diff}$}

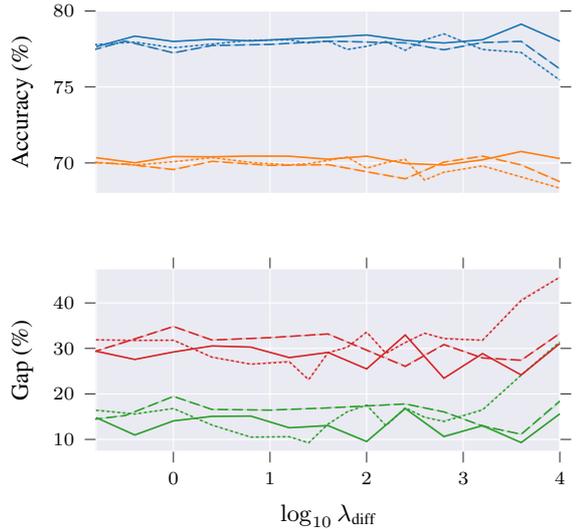
\begin{figure}[t!]
    \centering
    \pgfplotsset{width=0.48\textwidth, height = 4cm, compat=1.16}
    \pgfplotsset{every axis/.append style={
                    label style={font=\small},
                    tick label style={font=\scriptsize}  
                    }}
    \input{Figures/figure3}
    \caption{
      $\lambda_{\diff}$ sensitivity analysis for differentiated adversarial models with 3 (``\sampleline{}''), 5 (``\sampleline{dashed}''), and 8 (``\sampleline{dotted}'') sub-discriminators, in terms of the main task accuracy of group \textcolor{acc0}{\textit{SAE}} (blue) and \textcolor{acc1}{\textit{AAE}} (orange), and \textcolor{tpr_gap}{\textit{TPR-GAP}} (green) and \textcolor{tnr_gap}{\textit{TNR-GAP}} (red).
    }
    \label{fig:lambda_diff}
\end{figure}

Figure~\ref{fig:lambda_diff} presents the results of our  model with different $\lambda_{\diff}$ values, for $N \in \{3,5,8\}$ sub-discriminators. 

\final{
First, note that when $\lambda_{\diff}$ is small (i.e., the left side of Figure~\ref{fig:lambda_diff}), our Differentiated Adv Ensemble model generalizes to the standard Adv Ensemble model.
For differing numbers of sub-discriminators, performance is similar, i.e., increasing the number of sub-discriminators beyond 3 does not improve results substantially, but does come with a computational cost. 
This implies that an Adv Ensemble model learns approximately the same thing as larger ensembles (but more efficiently), where the sub-discriminators can only be explicitly differentiated by their weight initializations (with different random seeds), noting that all sub-discriminators are otherwise identical in architecture, input, and optimizer.
}

Increasing the weight of the difference loss through $\lambda_{\diff}$ has a positive influence on results, but an overly large value makes the sub-discriminators underfit, and both reduces accuracy and increases TPR/TNR Gap.
We observe a negative correlation between $N$ and $\lambda_{\diff}$, the main reason being that $\mathcal{L}_{\diff}$ is not averaged over $N$ and as a result, a large $N$ and $\lambda_{\diff}$ force the sub-discriminators to pay too much attention to orthogonality, impeding their ability to bleach out the protected attributes.

\final{
Overall, we empirically show that $\lambda_{\diff}$ only needs to be tuned for Adv Ensemble, since the results for different Differentiated Adv models for a given setting achieve similar results. I.e., $\lambda_{\diff}$ can safely be tuned separately with all other hyperparameters fixed.
}

\section{Conclusion and Future Work}
We have proposed an approach to enhance sub-discriminators in adversarial ensembles by introducing a difference loss.
Over a tweet sentiment classification task, we showed that our method substantially improves over standard adversarial methods, including ensemble-based methods.

\final{
In future work, we intend to perform experimentation over other tasks. Theoretically, our approach is general-purpose, and can be used not only for adversarial debiasing but also any other application where adversarial training is used, such as domain adaptation \citep{li-etal-2018-whats}.  
% We are also interested in developing effective frameworks for tuning hyperparameters more readily, such that our Differentiated Ensemble method can be readily deployed across a broad range of deep learning problems.
}

\section*{Acknowledgments}
We thank Lea Frermann, Shivashankar Subramanian, and the anonymous reviewers for their helpful feedback and suggestions.

% \bibliography{anthology,eacl2021}
\bibliography{eacl2021}
\bibliographystyle{acl_natbib}

% \appendix

\end{document}

%% file: Figures/figure2.tex
% This file was created by tikzplotlib v0.9.6.
\begin{tikzpicture}

\definecolor{color0}{rgb}{0.917647058823529,0.917647058823529,0.949019607843137}
\definecolor{color1}{rgb}{0.12156862745098,0.466666666666667,0.705882352941177}
\definecolor{color2}{rgb}{1,0.498039215686275,0.0549019607843137}
\definecolor{color3}{rgb}{0.172549019607843,0.627450980392157,0.172549019607843}
\definecolor{color4}{rgb}{0.83921568627451,0.152941176470588,0.156862745098039}

\begin{groupplot}[group style={group size=1 by 2}]
\nextgroupplot[
axis background/.style={fill=color0},
axis line style={white},
legend cell align={left},
legend style={fill opacity=0.8, draw opacity=1, text opacity=1, draw=white!80!black, fill=color0},
scaled x ticks=manual:{}{\pgfmathparse{#1}},
tick align=outside,
x grid style={white},
xlabel={},
xmajorgrids,
xmajorticks=false,
xmin=-2, xmax=2,
xtick style={color=white!15!black},
xticklabels={},
y grid style={white},
ylabel={Accuracy (\%)},
ymajorgrids,
% ymajorticks=false,
ymin=52, ymax=80,
ytick style={color=white!15!black}
]
\path [draw=color1, fill=color1, opacity=0.2]
(axis cs:-3,75.5)
--(axis cs:-3,75.5)
--(axis cs:-2.8,75.495)
--(axis cs:-2.6,75.46)
--(axis cs:-2.4,75.61)
--(axis cs:-2.2,75.695)
--(axis cs:-2,75.675)
--(axis cs:-1.8,75.73)
--(axis cs:-1.6,75.755)
--(axis cs:-1.4,75.915)
--(axis cs:-1.2,76.07)
--(axis cs:-1,76.515)
--(axis cs:-0.8,77.225)
--(axis cs:-0.6,77.295)
--(axis cs:-0.4,77.81)
--(axis cs:-0.2,78.45)
--(axis cs:0,78.445)
--(axis cs:0.2,76.8475)
--(axis cs:0.4,71.745)
--(axis cs:0.6,71.695)
--(axis cs:0.8,58.775)
--(axis cs:1,60.47)
--(axis cs:1.2,57.515)
--(axis cs:1.4,62.81)
--(axis cs:1.6,62.86)
--(axis cs:1.8,60.735)
--(axis cs:2,59.685)
--(axis cs:2.2,60.03)
--(axis cs:2.4,57.16)
--(axis cs:2.6,55.905)
--(axis cs:2.8,57.74)
--(axis cs:2.8,57.74)
--(axis cs:2.8,57.74)
--(axis cs:2.6,55.905)
--(axis cs:2.4,57.16)
--(axis cs:2.2,60.03)
--(axis cs:2,59.685)
--(axis cs:1.8,60.735)
--(axis cs:1.6,62.86)
--(axis cs:1.4,62.81)
--(axis cs:1.2,57.515)
--(axis cs:1,60.47)
--(axis cs:0.8,58.775)
--(axis cs:0.6,71.695)
--(axis cs:0.4,71.745)
--(axis cs:0.2,76.8475)
--(axis cs:0,78.445)
--(axis cs:-0.2,78.45)
--(axis cs:-0.4,77.81)
--(axis cs:-0.6,77.295)
--(axis cs:-0.8,77.225)
--(axis cs:-1,76.515)
--(axis cs:-1.2,76.07)
--(axis cs:-1.4,75.915)
--(axis cs:-1.6,75.755)
--(axis cs:-1.8,75.73)
--(axis cs:-2,75.675)
--(axis cs:-2.2,75.695)
--(axis cs:-2.4,75.61)
--(axis cs:-2.6,75.46)
--(axis cs:-2.8,75.495)
--(axis cs:-3,75.5)
--cycle;

\path [draw=color2, fill=color2, opacity=0.2]
(axis cs:-3,68.4392196098049)
--(axis cs:-3,68.4392196098049)
--(axis cs:-2.8,68.4192096048024)
--(axis cs:-2.6,68.5392696348174)
--(axis cs:-2.4,68.4792396198099)
--(axis cs:-2.2,68.4792396198099)
--(axis cs:-2,68.5292646323162)
--(axis cs:-1.8,68.4492246123062)
--(axis cs:-1.6,68.624312156078)
--(axis cs:-1.4,68.5592796398199)
--(axis cs:-1.2,68.6943471735868)
--(axis cs:-1,68.744372186093)
--(axis cs:-0.8,68.911955977989)
--(axis cs:-0.6,69.1395697848925)
--(axis cs:-0.4,69.6498249124562)
--(axis cs:-0.2,70.0750375187594)
--(axis cs:0,70.0200100050025)
--(axis cs:0.2,69.8299149574787)
--(axis cs:0.4,64.0120060030015)
--(axis cs:0.6,63.751875937969)
--(axis cs:0.8,53.6618309154577)
--(axis cs:1,56.408204102051)
--(axis cs:1.2,54.472236118059)
--(axis cs:1.4,58.8744372186093)
--(axis cs:1.6,59.5247623811906)
--(axis cs:1.8,55.8429214607304)
--(axis cs:2,56.7383691845923)
--(axis cs:2.2,60.32016008004)
--(axis cs:2.4,55.7778889444722)
--(axis cs:2.6,56.0005002501251)
--(axis cs:2.8,58.664332166083)
--(axis cs:2.8,58.664332166083)
--(axis cs:2.8,58.664332166083)
--(axis cs:2.6,56.0005002501251)
--(axis cs:2.4,55.7778889444722)
--(axis cs:2.2,60.32016008004)
--(axis cs:2,56.7383691845923)
--(axis cs:1.8,55.8429214607304)
--(axis cs:1.6,59.5247623811906)
--(axis cs:1.4,58.8744372186093)
--(axis cs:1.2,54.472236118059)
--(axis cs:1,56.408204102051)
--(axis cs:0.8,53.6618309154577)
--(axis cs:0.6,63.751875937969)
--(axis cs:0.4,64.0120060030015)
--(axis cs:0.2,69.8299149574787)
--(axis cs:0,70.0200100050025)
--(axis cs:-0.2,70.0750375187594)
--(axis cs:-0.4,69.6498249124562)
--(axis cs:-0.6,69.1395697848925)
--(axis cs:-0.8,68.911955977989)
--(axis cs:-1,68.744372186093)
--(axis cs:-1.2,68.6943471735868)
--(axis cs:-1.4,68.5592796398199)
--(axis cs:-1.6,68.624312156078)
--(axis cs:-1.8,68.4492246123062)
--(axis cs:-2,68.5292646323162)
--(axis cs:-2.2,68.4792396198099)
--(axis cs:-2.4,68.4792396198099)
--(axis cs:-2.6,68.5392696348174)
--(axis cs:-2.8,68.4192096048024)
--(axis cs:-3,68.4392196098049)
--cycle;

\addplot [semithick, color1, dash pattern=on 1pt off 3pt on 3pt off 3pt, forget plot]
table {%
-2.20000004768372 75.6900024414062
-2 75.6750030517578
-1.79999995231628 75.7300033569336
-1.60000002384186 75.754997253418
-1.39999997615814 75.9150009155273
-1.20000004768372 76.0699996948242
-1 76.5149993896484
-0.799999952316284 77.2249984741211
-0.600000023841858 77.2750015258789
-0.399999976158142 77.8099975585938
-0.200000047683716 78.4499969482422
0 78.4449996948242
0.200000047683716 76.8450012207031
0.399999976158142 71.7600021362305
0.600000023841858 71.6949996948242
0.799999952316284 58.7750015258789
1 60.4700012207031
1.20000004768372 57.3800010681152
1.39999997615814 62.810001373291
1.60000002384186 62.8600006103516
1.79999995231628 60.7350006103516
2 59.685001373291
};
\addplot [semithick, color2, dash pattern=on 1pt off 3pt on 3pt off 3pt, forget plot]
table {%
-2.20000004768372 68.4792404174805
-2 68.5292663574219
-1.79999995231628 68.444221496582
-1.60000002384186 68.6243133544922
-1.39999997615814 68.5592803955078
-1.20000004768372 68.6943435668945
-1 68.7443695068359
-0.799999952316284 68.9094543457031
-0.600000023841858 69.1395721435547
-0.399999976158142 69.6498260498047
-0.200000047683716 70.0650329589844
0 70.015007019043
0.200000047683716 69.7748870849609
0.399999976158142 64.0120086669922
0.600000023841858 63.7518768310547
0.799999952316284 53.8019027709961
1 56.408203125
1.20000004768372 54.6223106384277
1.39999997615814 58.8744354248047
1.60000002384186 59.5247611999512
1.79999995231628 55.9229621887207
2 56.4732360839844
};

\nextgroupplot[
axis background/.style={fill=color0},
axis line style={white},
legend cell align={left},
legend style={fill opacity=0.8, draw opacity=1, text opacity=1, draw=white!80!black, fill=color0},
tick align=outside,
x grid style={white},
xlabel={\(\displaystyle \log_{10}\lambda_{\adv}\)},
xmajorgrids,
% xmajorticks=false,
xmin=-2, xmax=2,
xtick style={color=white!15!black},
y grid style={white},
ylabel={GAP (\%)},
ymajorgrids,
% ymajorticks=false,
ymin=-0.543720860430213, ymax=48.297907953977,
ytick style={color=white!15!black}
]
\path [draw=color3, fill=color3, opacity=0.2]
(axis cs:-3,31.8463331665833)
--(axis cs:-3,31.8463331665833)
--(axis cs:-2.8,31.9362331165583)
--(axis cs:-2.6,32.1162931465733)
--(axis cs:-2.4,31.7463931965983)
--(axis cs:-2.2,31.6363581790896)
--(axis cs:-2,31.5863731865933)
--(axis cs:-1.8,31.3262931465733)
--(axis cs:-1.6,31.156288144072)
--(axis cs:-1.4,30.6762181090545)
--(axis cs:-1.2,30.0360780390195)
--(axis cs:-1,28.485947973987)
--(axis cs:-0.8,26.3259979989995)
--(axis cs:-0.6,24.8854927463732)
--(axis cs:-0.4,22.2750075037519)
--(axis cs:-0.2,15.0031465732866)
--(axis cs:0,10.8239669834917)
--(axis cs:0.2,6.52016008004002)
--(axis cs:0.4,17.3104252126063)
--(axis cs:0.6,13.6540070035018)
--(axis cs:0.8,7.33150075037519)
--(axis cs:1,11.166308154077)
--(axis cs:1.2,3.86098549274637)
--(axis cs:1.4,18.9261930965483)
--(axis cs:1.6,11.2291695847924)
--(axis cs:1.8,19.2632491245623)
--(axis cs:2,12.4956728364182)
--(axis cs:2.2,24.3105752876438)
--(axis cs:2.4,16.3217258629315)
--(axis cs:2.6,13.2525262631316)
--(axis cs:2.8,18.315947973987)
--(axis cs:2.8,18.315947973987)
--(axis cs:2.8,18.315947973987)
--(axis cs:2.6,13.2525262631316)
--(axis cs:2.4,16.3217258629315)
--(axis cs:2.2,24.3105752876438)
--(axis cs:2,12.4956728364182)
--(axis cs:1.8,19.2632491245623)
--(axis cs:1.6,11.2291695847924)
--(axis cs:1.4,18.9261930965483)
--(axis cs:1.2,3.86098549274637)
--(axis cs:1,11.166308154077)
--(axis cs:0.8,7.33150075037519)
--(axis cs:0.6,13.6540070035018)
--(axis cs:0.4,17.3104252126063)
--(axis cs:0.2,6.52016008004002)
--(axis cs:0,10.8239669834917)
--(axis cs:-0.2,15.0031465732866)
--(axis cs:-0.4,22.2750075037519)
--(axis cs:-0.6,24.8854927463732)
--(axis cs:-0.8,26.3259979989995)
--(axis cs:-1,28.485947973987)
--(axis cs:-1.2,30.0360780390195)
--(axis cs:-1.4,30.6762181090545)
--(axis cs:-1.6,31.156288144072)
--(axis cs:-1.8,31.3262931465733)
--(axis cs:-2,31.5863731865933)
--(axis cs:-2.2,31.6363581790896)
--(axis cs:-2.4,31.7463931965983)
--(axis cs:-2.6,32.1162931465733)
--(axis cs:-2.8,31.9362331165583)
--(axis cs:-3,31.8463331665833)
--cycle;

\path [draw=color4, fill=color4, opacity=0.2]
(axis cs:-3,45.9678939469735)
--(axis cs:-3,45.9678939469735)
--(axis cs:-2.8,46.077803901951)
--(axis cs:-2.6,45.9477538769385)
--(axis cs:-2.4,46.0079139569785)
--(axis cs:-2.2,46.0679139569785)
--(axis cs:-2,45.887823911956)
--(axis cs:-1.8,45.8978689344672)
--(axis cs:-1.6,45.417663831916)
--(axis cs:-1.4,45.3876738369185)
--(axis cs:-1.2,44.7873836918459)
--(axis cs:-1,44.0272036018009)
--(axis cs:-0.8,42.9570885442721)
--(axis cs:-0.6,41.1463031515758)
--(axis cs:-0.4,38.5953576788394)
--(axis cs:-0.2,31.7431965982992)
--(axis cs:0,27.6231465732866)
--(axis cs:0.2,16.6296298149075)
--(axis cs:0.4,6.50196098049024)
--(axis cs:0.6,6.13539269634817)
--(axis cs:0.8,4.37768384192096)
--(axis cs:1,5.8672236118059)
--(axis cs:1.2,1.6763531765883)
--(axis cs:1.4,27.2046323161581)
--(axis cs:1.6,14.9291445722861)
--(axis cs:1.8,28.5379739869935)
--(axis cs:2,18.9191995997999)
--(axis cs:2.2,20.1720060030015)
--(axis cs:2.4,19.045947973987)
--(axis cs:2.6,9.55697848924462)
--(axis cs:2.8,16.8846723361681)
--(axis cs:2.8,16.8846723361681)
--(axis cs:2.8,16.8846723361681)
--(axis cs:2.6,9.55697848924462)
--(axis cs:2.4,19.045947973987)
--(axis cs:2.2,20.1720060030015)
--(axis cs:2,18.9191995997999)
--(axis cs:1.8,28.5379739869935)
--(axis cs:1.6,14.9291445722861)
--(axis cs:1.4,27.2046323161581)
--(axis cs:1.2,1.6763531765883)
--(axis cs:1,5.8672236118059)
--(axis cs:0.8,4.37768384192096)
--(axis cs:0.6,6.13539269634817)
--(axis cs:0.4,6.50196098049024)
--(axis cs:0.2,16.6296298149075)
--(axis cs:0,27.6231465732866)
--(axis cs:-0.2,31.7431965982992)
--(axis cs:-0.4,38.5953576788394)
--(axis cs:-0.6,41.1463031515758)
--(axis cs:-0.8,42.9570885442721)
--(axis cs:-1,44.0272036018009)
--(axis cs:-1.2,44.7873836918459)
--(axis cs:-1.4,45.3876738369185)
--(axis cs:-1.6,45.417663831916)
--(axis cs:-1.8,45.8978689344672)
--(axis cs:-2,45.887823911956)
--(axis cs:-2.2,46.0679139569785)
--(axis cs:-2.4,46.0079139569785)
--(axis cs:-2.6,45.9477538769385)
--(axis cs:-2.8,46.077803901951)
--(axis cs:-3,45.9678939469735)
--cycle;

\addplot [semithick, color3, dash pattern=on 1pt off 3pt on 3pt off 3pt, forget plot]
table {%
-2 31.5963535308838
-1.79999995231628 31.3262672424316
-1.60000002384186 31.1562881469727
-1.39999997615814 30.6762237548828
-1.20000004768372 30.0360774993896
-1 28.4859485626221
-0.799999952316284 26.3259983062744
-0.600000023841858 24.8854923248291
-0.399999976158142 22.2750072479248
-0.200000047683716 15.0733366012573
0.200000047683716 6.460205078125
0.399999976158142 16.7720966339111
0.600000023841858 13.6540069580078
0.799999952316284 7.33689832687378
1 11.0564527511597
1.20000004768372 3.95896458625793
1.39999997615814 18.9261932373047
1.60000002384186 11.5778188705444
1.79999995231628 19.5253772735596
2 12.4956731796265
};
\addplot [semithick, color4, dash pattern=on 1pt off 3pt on 3pt off 3pt, forget plot]
table {%
-2 45.887825012207
-1.79999995231628 45.8978233337402
-1.60000002384186 45.4176635742188
-1.39999997615814 45.3876647949219
-1.20000004768372 44.7873840332031
-1 44.0272026062012
-0.799999952316284 42.957088470459
-0.600000023841858 41.1563529968262
-0.399999976158142 38.5953559875488
-0.200000047683716 31.8432712554932
0 27.6231460571289
0.200000047683716 16.6004295349121
0.399999976158142 6.59696865081787
0.600000023841858 6.13539266586304
0.799999952316284 4.37768363952637
1 5.89694356918335
1.20000004768372 1.67635321617126
1.39999997615814 26.7973194122314
1.60000002384186 14.950945854187
1.79999995231628 28.4760589599609
2 18.9191989898682
};
\end{groupplot}

\end{tikzpicture}

%% file: Figures/figure3.tex
% This file was created by tikzplotlib v0.9.6.
\begin{tikzpicture}

\definecolor{color0}{rgb}{0.917647058823529,0.917647058823529,0.949019607843137}
\definecolor{color1}{rgb}{0.12156862745098,0.466666666666667,0.705882352941177}
\definecolor{color2}{rgb}{1,0.498039215686275,0.0549019607843137}
\definecolor{color3}{rgb}{0.172549019607843,0.627450980392157,0.172549019607843}
\definecolor{color4}{rgb}{0.83921568627451,0.152941176470588,0.156862745098039}

\begin{groupplot}[group style={group size=1 by 2}]
\nextgroupplot[
axis background/.style={fill=color0},
axis line style={white},
legend style={fill opacity=0.8, draw opacity=1, text opacity=1, draw=none, fill=color0},
scaled x ticks=manual:{}{\pgfmathparse{#1}},
tick align=outside,
x grid style={white},
xlabel={},
xmajorgrids,
xmajorticks=false,
xmin=-0.8, xmax=4,
xtick style={color=white!15!black},
xticklabels={},
y grid style={white},
ylabel={Accuracy (\%)},
ymajorgrids,
% ymajorticks=false,
ymin=68, ymax=80,
ytick style={color=white!15!black}
]
\path [draw=color1, fill=color1, opacity=0.2]
(axis cs:-0.8,77.65)
--(axis cs:-0.8,77.65)
--(axis cs:-0.4,78.345)
--(axis cs:0,77.9975)
--(axis cs:0.4,78.14)
--(axis cs:0.8,78.02)
--(axis cs:1.2,78.16)
--(axis cs:1.6,78.2775)
--(axis cs:2,78.4225)
--(axis cs:2.4,78.09)
--(axis cs:2.8,77.885)
--(axis cs:3.2,78.105)
--(axis cs:3.6,79.1325)
--(axis cs:4,78.01)
--(axis cs:4,78.01)
--(axis cs:4,78.01)
--(axis cs:3.6,79.1325)
--(axis cs:3.2,78.105)
--(axis cs:2.8,77.885)
--(axis cs:2.4,78.09)
--(axis cs:2,78.4225)
--(axis cs:1.6,78.2775)
--(axis cs:1.2,78.16)
--(axis cs:0.8,78.02)
--(axis cs:0.4,78.14)
--(axis cs:0,77.9975)
--(axis cs:-0.4,78.345)
--(axis cs:-0.8,77.65)
--cycle;

\path [draw=color1, fill=color1, opacity=0.2]
(axis cs:-1,77.14)
--(axis cs:-1,77.14)
--(axis cs:-0.5,78.02)
--(axis cs:0,77.245)
--(axis cs:0.4,77.73)
--(axis cs:1,77.8)
--(axis cs:1.6,78.015)
--(axis cs:2.4,77.9)
--(axis cs:2.8,77.45)
--(axis cs:3.2,77.93)
--(axis cs:3.6,78)
--(axis cs:4,76.205)
--(axis cs:4,76.205)
--(axis cs:4,76.205)
--(axis cs:3.6,78)
--(axis cs:3.2,77.93)
--(axis cs:2.8,77.45)
--(axis cs:2.4,77.9)
--(axis cs:1.6,78.015)
--(axis cs:1,77.8)
--(axis cs:0.4,77.73)
--(axis cs:0,77.245)
--(axis cs:-0.5,78.02)
--(axis cs:-1,77.14)
--cycle;

\path [draw=color1, fill=color1, opacity=0.2]
(axis cs:-2,77.89)
--(axis cs:-2,77.89)
--(axis cs:-1.6,78.185)
--(axis cs:-1.2,77.395)
--(axis cs:-0.8,77.79)
--(axis cs:-0.4,77.955)
--(axis cs:0,77.58)
--(axis cs:0.4,77.845)
--(axis cs:0.8,78.075)
--(axis cs:1.2,78.1)
--(axis cs:1.4,77.895)
--(axis cs:1.6,78.02)
--(axis cs:1.8,77.47)
--(axis cs:2,77.695)
--(axis cs:2.2,77.985)
--(axis cs:2.4,77.415)
--(axis cs:2.6,78.095)
--(axis cs:2.8,78.4975)
--(axis cs:3.2,77.465)
--(axis cs:3.6,77.28625)
--(axis cs:4,75.4575)
--(axis cs:4,75.4575)
--(axis cs:4,75.4575)
--(axis cs:3.6,77.28625)
--(axis cs:3.2,77.465)
--(axis cs:2.8,78.4975)
--(axis cs:2.6,78.095)
--(axis cs:2.4,77.415)
--(axis cs:2.2,77.985)
--(axis cs:2,77.695)
--(axis cs:1.8,77.47)
--(axis cs:1.6,78.02)
--(axis cs:1.4,77.895)
--(axis cs:1.2,78.1)
--(axis cs:0.8,78.075)
--(axis cs:0.4,77.845)
--(axis cs:0,77.58)
--(axis cs:-0.4,77.955)
--(axis cs:-0.8,77.79)
--(axis cs:-1.2,77.395)
--(axis cs:-1.6,78.185)
--(axis cs:-2,77.89)
--cycle;

\path [draw=color2, fill=color2, opacity=0.2]
(axis cs:-0.8,70.360180090045)
--(axis cs:-0.8,70.360180090045)
--(axis cs:-0.4,70.0050025012506)
--(axis cs:0,70.4202101050525)
--(axis cs:0.4,70.4052026013007)
--(axis cs:0.8,70.4602301150575)
--(axis cs:1.2,70.4427213606803)
--(axis cs:1.6,70.2651325662831)
--(axis cs:2,70.4502251125563)
--(axis cs:2.4,69.9699849924963)
--(axis cs:2.8,69.8699349674837)
--(axis cs:3.2,70.2251125562781)
--(axis cs:3.6,70.760380190095)
--(axis cs:4,70.2951475737869)
--(axis cs:4,70.2951475737869)
--(axis cs:4,70.2951475737869)
--(axis cs:3.6,70.760380190095)
--(axis cs:3.2,70.2251125562781)
--(axis cs:2.8,69.8699349674837)
--(axis cs:2.4,69.9699849924963)
--(axis cs:2,70.4502251125563)
--(axis cs:1.6,70.2651325662831)
--(axis cs:1.2,70.4427213606803)
--(axis cs:0.8,70.4602301150575)
--(axis cs:0.4,70.4052026013007)
--(axis cs:0,70.4202101050525)
--(axis cs:-0.4,70.0050025012506)
--(axis cs:-0.8,70.360180090045)
--cycle;

\path [draw=color2, fill=color2, opacity=0.2]
(axis cs:-1,70.1150575287644)
--(axis cs:-1,70.1150575287644)
--(axis cs:-0.5,69.9249624812406)
--(axis cs:0,69.5747873936969)
--(axis cs:0.4,70.1100550275138)
--(axis cs:1,69.83991995998)
--(axis cs:1.6,69.8899449724862)
--(axis cs:2.4,68.9844922461231)
--(axis cs:2.8,70.0550275137569)
--(axis cs:3.2,70.4602301150575)
--(axis cs:3.6,69.8999499749875)
--(axis cs:4,68.7993996998499)
--(axis cs:4,68.7993996998499)
--(axis cs:4,68.7993996998499)
--(axis cs:3.6,69.8999499749875)
--(axis cs:3.2,70.4602301150575)
--(axis cs:2.8,70.0550275137569)
--(axis cs:2.4,68.9844922461231)
--(axis cs:1.6,69.8899449724862)
--(axis cs:1,69.83991995998)
--(axis cs:0.4,70.1100550275138)
--(axis cs:0,69.5747873936969)
--(axis cs:-0.5,69.9249624812406)
--(axis cs:-1,70.1150575287644)
--cycle;

\path [draw=color2, fill=color2, opacity=0.2]
(axis cs:-2,69.9049524762381)
--(axis cs:-2,69.9049524762381)
--(axis cs:-1.6,70.1750875437719)
--(axis cs:-1.2,69.9399699849925)
--(axis cs:-0.8,70.0450225112556)
--(axis cs:-0.4,69.879939969985)
--(axis cs:0,70.0950475237619)
--(axis cs:0.4,70.3551775887944)
--(axis cs:0.8,70.0350175087544)
--(axis cs:1.2,69.8749374687344)
--(axis cs:1.4,69.9549774887444)
--(axis cs:1.6,70.1750875437719)
--(axis cs:1.8,70.3801900950475)
--(axis cs:2,69.6548274137068)
--(axis cs:2.2,70.0450225112556)
--(axis cs:2.4,70.2451225612806)
--(axis cs:2.6,68.9044522261131)
--(axis cs:2.8,69.3896948474237)
--(axis cs:3.2,69.8149074537269)
--(axis cs:3.6,69.0795397698849)
--(axis cs:4,68.3491745872936)
--(axis cs:4,68.3491745872936)
--(axis cs:4,68.3491745872936)
--(axis cs:3.6,69.0795397698849)
--(axis cs:3.2,69.8149074537269)
--(axis cs:2.8,69.3896948474237)
--(axis cs:2.6,68.9044522261131)
--(axis cs:2.4,70.2451225612806)
--(axis cs:2.2,70.0450225112556)
--(axis cs:2,69.6548274137068)
--(axis cs:1.8,70.3801900950475)
--(axis cs:1.6,70.1750875437719)
--(axis cs:1.4,69.9549774887444)
--(axis cs:1.2,69.8749374687344)
--(axis cs:0.8,70.0350175087544)
--(axis cs:0.4,70.3551775887944)
--(axis cs:0,70.0950475237619)
--(axis cs:-0.4,69.879939969985)
--(axis cs:-0.8,70.0450225112556)
--(axis cs:-1.2,69.9399699849925)
--(axis cs:-1.6,70.1750875437719)
--(axis cs:-2,69.9049524762381)
--cycle;

\addplot [semithick, color1, forget plot]
table {%
-0.799999952316284 77.6500015258789
-0.399999976158142 78.3450012207031
0 77.995002746582
0.399999976158142 78.1399993896484
0.799999952316284 78.0299987792969
1.20000004768372 78.1600036621094
1.60000002384186 78.2699966430664
2 78.4199981689453
2.40000009536743 78.0599975585938
2.79999995231628 77.8949966430664
3.20000004768372 78.1050033569336
3.59999990463257 79.1350021362305
4 78.0100021362305
};
\addplot [semithick, color1, dash pattern=on 4pt off 1.5pt, forget plot]
table {%
-1 77.1399993896484
-0.5 78.0199966430664
0 77.245002746582
0.399999976158142 77.7300033569336
1 77.8000030517578
1.60000002384186 78.0149993896484
2.40000009536743 77.9000015258789
2.79999995231628 77.4449996948242
3.20000004768372 77.9250030517578
3.59999990463257 78
4 76.1849975585938
};
\addplot [semithick, color1, dash pattern=on 1pt off 1pt, forget plot]
table {%
-1.20000004768372 77.3949966430664
-0.799999952316284 77.7900009155273
-0.399999976158142 77.9550018310547
0 77.5800018310547
0.399999976158142 77.8150024414062
0.799999952316284 78.0749969482422
1.20000004768372 78.0999984741211
1.39999997615814 77.8949966430664
1.60000002384186 78.0199966430664
1.79999995231628 77.4700012207031
2 77.6699981689453
2.20000004768372 77.9850006103516
2.40000009536743 77.4150009155273
2.59999990463257 78.0950012207031
2.79999995231628 78.495002746582
3.20000004768372 77.4649963378906
3.59999990463257 77.2649993896484
4 75.4574966430664
};
\addplot [semithick, color2, forget plot]
table {%
-0.799999952316284 70.3451690673828
-0.399999976158142 70.0100021362305
0 70.4252090454102
0.399999976158142 70.4052047729492
0.799999952316284 70.4552307128906
1.20000004768372 70.4452209472656
1.60000002384186 70.2451248168945
2 70.4502258300781
2.40000009536743 69.9699859619141
2.79999995231628 69.8649291992188
3.20000004768372 70.2051010131836
3.59999990463257 70.7553787231445
4 70.2901458740234
};
\addplot [semithick, color2, dash pattern=on 4pt off 1.5pt, forget plot]
table {%
-1 70.0700378417969
-0.5 69.9299621582031
0 69.5647811889648
0.399999976158142 70.1050491333008
1 69.8399200439453
1.60000002384186 69.8899459838867
2.40000009536743 68.954475402832
2.79999995231628 70.0550308227539
3.20000004768372 70.4502258300781
3.59999990463257 69.8699340820312
4 68.7643814086914
};
\addplot [semithick, color2, dash pattern=on 1pt off 1pt, forget plot]
table {%
-1.20000004768372 69.9449691772461
-0.799999952316284 70.0450210571289
-0.399999976158142 69.8649291992188
0 70.0750350952148
0.399999976158142 70.3551788330078
0.799999952316284 70.0350189208984
1.20000004768372 69.8749389648438
1.39999997615814 69.9549789428711
1.60000002384186 70.1750869750977
1.79999995231628 70.3801879882812
2 69.6548309326172
2.20000004768372 70.0200119018555
2.40000009536743 70.240119934082
2.59999990463257 68.8644332885742
2.79999995231628 69.3946990966797
3.20000004768372 69.8049011230469
3.59999990463257 69.0795364379883
4 68.341667175293
};

\nextgroupplot[
axis background/.style={fill=color0},
axis line style={white},
legend cell align={left},
legend columns=3,
legend style={fill opacity=0.8, draw opacity=1, text opacity=1, at={(0.5,-0.2)}, anchor=north, draw=white!80!black, fill=color0},
tick align=outside,
x grid style={white},
xlabel={\(\displaystyle \log_{10}\lambda_{\diff}\)},
xmajorgrids,
% xmajorticks=false,
xmin=-0.8, xmax=4,
xtick style={color=white!15!black},
y grid style={white},
ylabel={Gap (\%)},
ymajorgrids,
% ymajorticks=false,
ymin=7.39870635317659, ymax=47.4796028014007,
ytick style={color=white!15!black}
]
\path [draw=color3, fill=color3, opacity=0.2]
(axis cs:-0.8,14.7923911955978)
--(axis cs:-0.8,14.7923911955978)
--(axis cs:-0.4,11.0863631815908)
--(axis cs:0,14.0728514257129)
--(axis cs:0.4,15.0730965482741)
--(axis cs:0.8,15.1329764882441)
--(axis cs:1.2,12.5215057528764)
--(axis cs:1.6,12.9133566783392)
--(axis cs:2,9.44154577288644)
--(axis cs:2.4,16.784132066033)
--(axis cs:2.8,10.8584542271136)
--(axis cs:3.2,12.8228064032016)
--(axis cs:3.6,9.2648024012006)
--(axis cs:4,15.6231065532766)
--(axis cs:4,15.6231065532766)
--(axis cs:4,15.6231065532766)
--(axis cs:3.6,9.2648024012006)
--(axis cs:3.2,12.8228064032016)
--(axis cs:2.8,10.8584542271136)
--(axis cs:2.4,16.784132066033)
--(axis cs:2,9.44154577288644)
--(axis cs:1.6,12.9133566783392)
--(axis cs:1.2,12.5215057528764)
--(axis cs:0.8,15.1329764882441)
--(axis cs:0.4,15.0730965482741)
--(axis cs:0,14.0728514257129)
--(axis cs:-0.4,11.0863631815908)
--(axis cs:-0.8,14.7923911955978)
--cycle;

\path [draw=color3, fill=color3, opacity=0.2]
(axis cs:-1,13.9624662331166)
--(axis cs:-1,13.9624662331166)
--(axis cs:-0.5,15.243731865933)
--(axis cs:0,19.4540020010005)
--(axis cs:0.4,16.5835617808904)
--(axis cs:1,16.443791895948)
--(axis cs:1.6,16.9437568784392)
--(axis cs:2.4,17.7900950475238)
--(axis cs:2.8,16.0532166083042)
--(axis cs:3.2,12.9527363681841)
--(axis cs:3.6,11.1328114057028)
--(axis cs:4,18.1243421710855)
--(axis cs:4,18.1243421710855)
--(axis cs:4,18.1243421710855)
--(axis cs:3.6,11.1328114057028)
--(axis cs:3.2,12.9527363681841)
--(axis cs:2.8,16.0532166083042)
--(axis cs:2.4,17.7900950475238)
--(axis cs:1.6,16.9437568784392)
--(axis cs:1,16.443791895948)
--(axis cs:0.4,16.5835617808904)
--(axis cs:0,19.4540020010005)
--(axis cs:-0.5,15.243731865933)
--(axis cs:-1,13.9624662331166)
--cycle;

\path [draw=color3, fill=color3, opacity=0.2]
(axis cs:-2,14.2633666833417)
--(axis cs:-2,14.2633666833417)
--(axis cs:-1.6,10.8227463731866)
--(axis cs:-1.2,15.0870285142571)
--(axis cs:-0.8,16.533871935968)
--(axis cs:-0.4,15.5536818409205)
--(axis cs:0,16.7933666833417)
--(axis cs:0.4,13.1528414207104)
--(axis cs:0.8,10.5168434217109)
--(axis cs:1.2,10.5934267133567)
--(axis cs:1.4,9.22056528264132)
--(axis cs:1.6,13.2536568284142)
--(axis cs:1.8,16.072396198099)
--(axis cs:2,17.5444322161081)
--(axis cs:2.2,13.1435317658829)
--(axis cs:2.4,16.9427913956978)
--(axis cs:2.6,14.8749574787394)
--(axis cs:2.8,14.0897623811906)
--(axis cs:3.2,16.6140370185093)
--(axis cs:3.6,24.1405690345173)
--(axis cs:4,31.4211030515258)
--(axis cs:4,31.4211030515258)
--(axis cs:4,31.4211030515258)
--(axis cs:3.6,24.1405690345173)
--(axis cs:3.2,16.6140370185093)
--(axis cs:2.8,14.0897623811906)
--(axis cs:2.6,14.8749574787394)
--(axis cs:2.4,16.9427913956978)
--(axis cs:2.2,13.1435317658829)
--(axis cs:2,17.5444322161081)
--(axis cs:1.8,16.072396198099)
--(axis cs:1.6,13.2536568284142)
--(axis cs:1.4,9.22056528264132)
--(axis cs:1.2,10.5934267133567)
--(axis cs:0.8,10.5168434217109)
--(axis cs:0.4,13.1528414207104)
--(axis cs:0,16.7933666833417)
--(axis cs:-0.4,15.5536818409205)
--(axis cs:-0.8,16.533871935968)
--(axis cs:-1.2,15.0870285142571)
--(axis cs:-1.6,10.8227463731866)
--(axis cs:-2,14.2633666833417)
--cycle;

\path [draw=color4, fill=color4, opacity=0.2]
(axis cs:-0.8,29.231655827914)
--(axis cs:-0.8,29.231655827914)
--(axis cs:-0.4,27.5721010505253)
--(axis cs:0,29.2124262131065)
--(axis cs:0.4,30.5426913456728)
--(axis cs:0.8,30.3077688844422)
--(axis cs:1.2,27.881655827914)
--(axis cs:1.6,29.1127113556778)
--(axis cs:2,25.4907803901951)
--(axis cs:2.4,32.8440020010005)
--(axis cs:2.8,23.4392996498249)
--(axis cs:3.2,28.8628264132066)
--(axis cs:3.6,24.151752126063)
--(axis cs:4,31.1132016008004)
--(axis cs:4,31.1132016008004)
--(axis cs:4,31.1132016008004)
--(axis cs:3.6,24.151752126063)
--(axis cs:3.2,28.8628264132066)
--(axis cs:2.8,23.4392996498249)
--(axis cs:2.4,32.8440020010005)
--(axis cs:2,25.4907803901951)
--(axis cs:1.6,29.1127113556778)
--(axis cs:1.2,27.881655827914)
--(axis cs:0.8,30.3077688844422)
--(axis cs:0.4,30.5426913456728)
--(axis cs:0,29.2124262131065)
--(axis cs:-0.4,27.5721010505253)
--(axis cs:-0.8,29.231655827914)
--cycle;

\path [draw=color4, fill=color4, opacity=0.2]
(axis cs:-1,28.102396198099)
--(axis cs:-1,28.102396198099)
--(axis cs:-0.5,31.2938569284642)
--(axis cs:0,34.8144372186093)
--(axis cs:0.4,31.8334567283642)
--(axis cs:1,32.363951975988)
--(axis cs:1.6,33.1739769884942)
--(axis cs:2.4,26.1917208604302)
--(axis cs:2.8,30.9031165582791)
--(axis cs:3.2,27.9022861430715)
--(axis cs:3.6,27.4423611805903)
--(axis cs:4,33.2850875437719)
--(axis cs:4,33.2850875437719)
--(axis cs:4,33.2850875437719)
--(axis cs:3.6,27.4423611805903)
--(axis cs:3.2,27.9022861430715)
--(axis cs:2.8,30.9031165582791)
--(axis cs:2.4,26.1917208604302)
--(axis cs:1.6,33.1739769884942)
--(axis cs:1,32.363951975988)
--(axis cs:0.4,31.8334567283642)
--(axis cs:0,34.8144372186093)
--(axis cs:-0.5,31.2938569284642)
--(axis cs:-1,28.102396198099)
--cycle;

\path [draw=color4, fill=color4, opacity=0.2]
(axis cs:-2,30.2434667333667)
--(axis cs:-2,30.2434667333667)
--(axis cs:-1.6,26.7832166083042)
--(axis cs:-1.2,29.2028314157079)
--(axis cs:-0.8,31.8733366683342)
--(axis cs:-0.4,31.7435317658829)
--(axis cs:0,31.8032916458229)
--(axis cs:0.4,28.0424762381191)
--(axis cs:0.8,26.5326663331666)
--(axis cs:1.2,27.2134717358679)
--(axis cs:1.4,23.0965007503752)
--(axis cs:1.6,28.9521710855428)
--(axis cs:1.8,30.252016008004)
--(axis cs:2,33.5845022511256)
--(axis cs:2.2,28.9231715857929)
--(axis cs:2.4,31.2925512756378)
--(axis cs:2.6,33.3360930465233)
--(axis cs:2.8,32.2657978989495)
--(axis cs:3.2,31.793711855928)
--(axis cs:3.6,40.7563169084542)
--(axis cs:4,45.647783891946)
--(axis cs:4,45.647783891946)
--(axis cs:4,45.647783891946)
--(axis cs:3.6,40.7563169084542)
--(axis cs:3.2,31.793711855928)
--(axis cs:2.8,32.2657978989495)
--(axis cs:2.6,33.3360930465233)
--(axis cs:2.4,31.2925512756378)
--(axis cs:2.2,28.9231715857929)
--(axis cs:2,33.5845022511256)
--(axis cs:1.8,30.252016008004)
--(axis cs:1.6,28.9521710855428)
--(axis cs:1.4,23.0965007503752)
--(axis cs:1.2,27.2134717358679)
--(axis cs:0.8,26.5326663331666)
--(axis cs:0.4,28.0424762381191)
--(axis cs:0,31.8032916458229)
--(axis cs:-0.4,31.7435317658829)
--(axis cs:-0.8,31.8733366683342)
--(axis cs:-1.2,29.2028314157079)
--(axis cs:-1.6,26.7832166083042)
--(axis cs:-2,30.2434667333667)
--cycle;

\addplot [semithick, color3, forget plot]
table {%
-0.799999952316284 14.7923908233643
-0.399999976158142 10.9671840667725
0 14.0728511810303
0.399999976158142 15.0730962753296
0.799999952316284 15.1329765319824
1.20000004768372 12.5618963241577
1.60000002384186 13.0629568099976
2 9.55123043060303
2.40000009536743 16.7841320037842
2.79999995231628 10.6065683364868
3.20000004768372 13.0630311965942
3.59999990463257 9.30743885040283
4 15.6231069564819
};
\addplot [semithick, color3, dash pattern=on 4pt off 1.5pt, forget plot]
table {%
-1 13.9624662399292
-0.5 15.2437314987183
0 19.4540023803711
0.399999976158142 16.5835609436035
1 16.4437923431396
1.60000002384186 16.9238662719727
2.40000009536743 17.7998905181885
2.79999995231628 16.0532169342041
3.20000004768372 12.9527359008789
3.59999990463257 11.1328115463257
4 18.4438514709473
};
\addplot [semithick, color3, dash pattern=on 1pt off 1pt, forget plot]
table {%
-0.799999952316284 16.3833808898926
-0.399999976158142 15.5936317443848
0 16.7933673858643
0.399999976158142 13.1528415679932
0.799999952316284 10.5168437957764
1.20000004768372 10.5934267044067
1.39999997615814 9.24649810791016
1.60000002384186 13.3728713989258
1.79999995231628 16.072395324707
2 17.5541572570801
2.20000004768372 12.9931917190552
2.40000009536743 16.9427909851074
2.59999990463257 14.8749570846558
2.79999995231628 13.9347772598267
3.20000004768372 16.4735164642334
3.59999990463257 24.1303424835205
4 31.4260864257812
};
\addplot [semithick, color4, forget plot]
table {%
-0.799999952316284 29.4020462036133
-0.399999976158142 27.5721015930176
0 29.2124271392822
0.399999976158142 30.5426921844482
0.799999952316284 30.2825222015381
1.20000004768372 27.9914512634277
1.60000002384186 29.1127109527588
2 25.490779876709
2.40000009536743 32.9641609191895
2.79999995231628 23.4393005371094
3.20000004768372 28.8628273010254
3.59999990463257 24.1666831970215
4 31.062816619873
};
\addplot [semithick, color4, dash pattern=on 4pt off 1.5pt, forget plot]
table {%
-1 28.1023960113525
-0.5 31.4238014221191
0 34.8144378662109
0.399999976158142 31.8334560394287
1 32.3639526367188
1.60000002384186 33.1739768981934
2.40000009536743 26.0413112640381
2.79999995231628 30.8331623077393
3.20000004768372 27.902286529541
3.59999990463257 27.3929405212402
4 33.2850875854492
};
\addplot [semithick, color4, dash pattern=on 1pt off 1pt, forget plot]
table {%
-0.799999952316284 31.8733367919922
-0.399999976158142 31.7737674713135
0 31.8032913208008
0.399999976158142 28.0724868774414
0.799999952316284 26.5326671600342
1.20000004768372 27.0435523986816
1.39999997615814 23.1217308044434
1.60000002384186 29.0626964569092
1.79999995231628 30.2520160675049
2 33.5845031738281
2.20000004768372 28.9231719970703
2.40000009536743 31.2925510406494
2.59999990463257 33.3360939025879
2.79999995231628 32.135383605957
3.20000004768372 31.7937126159668
3.59999990463257 40.5012626647949
4 45.6577453613281
};
\end{groupplot}

\end{tikzpicture}